\documentclass{article}
\usepackage{spconf,amsmath,graphicx}
\usepackage[utf8]{inputenc} %
\usepackage[T1]{fontenc}    %
\usepackage[hidelinks]{hyperref}       %
\usepackage{url}            %
\usepackage{booktabs}       %
\usepackage{amsfonts}       %
\usepackage{nicefrac}       %
\usepackage{microtype}      %
\usepackage{xcolor}         %
\usepackage{graphicx}

\usepackage{amsmath}
\usepackage{multirow}
\usepackage{wrapfig}
\usepackage{caption}
\usepackage{appendix}
\usepackage{float}
\usepackage{enumitem}
\usepackage[normalem]{ulem} %

\def\flexquad{\hskip0.5em \relax}

\renewcommand\paragraph[1]{\noindent\textbf{#1:}\flexquad}
\usepackage{xcolor}
\definecolor{class}{rgb}{1.0, 0.13, 0.32}
\definecolor{targetclass}{rgb}{0.0, 0.2, 0.6 }
\newcommand{\class}[1]{\textcolor{class}{\textbf{\textit{#1}}}}

\definecolor{promptcolor}{rgb}{0.0, 0.55, 0.55}
\definecolor{descriptioncolor}{rgb}{0.55, 0.0, 0.0}

\newcommand{\targetclass}[1]{\textcolor{targetclass}{\textbf{\textit{#1}}}}

\newcommand{\llmprompt}[1]{\textcolor{promptcolor}{\textbf{#1}}}

\newcommand{\llmdescription}[1]{\textcolor{descriptioncolor}{\textbf{#1}}}

\newcommand{\phanthomp}[1]{\phantom{\textbf{P:}  }}
\newcommand{\phanthomd}[1]{\phantom{\textbf{D:}  }}

\makeatletter
\def\blfootnote{\gdef\@thefnmark{}\@footnotetext}
\makeatother
\title{TAP: Targeted Prompting for task adaptive generation of textual training instances for visual classification}
\name{{M. Jehanzeb Mirza$^{\dagger1}$ \flexquad Leonid Karlinsky$^{2}$ \flexquad Wei Lin$^{1}$ \flexquad Horst Possegger$^{1}$ \flexquad Rogerio Feris$^{2}$ \flexquad Horst Bischof$^{1}$}}
\address{$^{1}$Institute of Computer Graphics and Vision, TU Graz, Austria.\\$^{2}$MIT-IBM Watson AI Lab, USA.}

\newcommand{\vis}{v}  %
\newcommand{\txt}{u}  %
\newcommand{\visEnc}{\vis} %
\newcommand{\txtEnc}{\txt} %

\newcommand{\txtEmb}{\txt} %

\newcommand{\cls}{c} %
\newcommand{\ClassNames}{C} %

\newcommand{\img}{x} %

\newcommand{\simil}{\cos} %

\newcommand{\likel}{l} %

\newcommand{\LSCE}{\mathcal{L}_\textsc{SCE}}

\newif\ifdraft
\draftfalse
\drafttrue %
\ifdraft
 \newcommand{\MK}[1]{{\color{magenta}{\bf MK: #1}}}

\else
 \newcommand{\MK}[1]{}
 
\fi

\begin{document}

\ninept
\maketitle
\begin{abstract}
Vision and Language Models (VLMs), such as CLIP, have enabled visual recognition of a potentially unlimited set of categories described by text prompts. However, for the best visual recognition performance, these models still require tuning to better fit the data distributions of the downstream tasks, in order to overcome the domain shift from the web-based pre-training data. Recently, it has been shown that it is possible to effectively tune VLMs without any paired data, and in particular to effectively improve VLMs visual recognition performance using text-only training data generated by Large Language Models (LLMs). In this paper, we dive deeper into this exciting text-only VLM training approach and explore ways it can be significantly further improved taking the specifics of the downstream task into account when sampling text data from LLMs. In particular, compared to the SOTA text-only VLM training approach, we demonstrate up to $8.4\%$ performance improvement in (cross) domain-specific adaptation, up to $8.7\%$ improvement in fine-grained recognition, and $3.1\%$ overall average improvement in zero-shot classification compared to strong baselines. 
\end{abstract}
\begin{keywords}
Vision and Language Models, VLM, text-only training, zero-shot visual classification
\end{keywords}
\blfootnote{$^\dagger$Correspondence: \tt\small{muhammad.mirza@icg.tugraz.at}}
\vspace{-1em}
\section{Introduction}
\vspace{-1em}
\label{sec:intro}
Vision and Language Models (VLMs) pre-trained on large-scale paired image-text datasets collected from the web (e.g. CLIP \cite{clip}) have made unprecedented progress in recent years, enabling open-vocabulary zero-shot recognition, closely rivaling closed-set (fixed classes set) pre-trained models. 
The typical \textit{`generalist'} dual-encoder VLM zero-shot approach for (downstream) image classification uses a collection of human-engineered prompts that are encoded by the VLM text-encoder and compared to the image in question encoded by the VLM image-encoder, determining the image's class by maximal average similarity. 
However, as noted by several works \cite{coop,cocoop}, for better performance, this VLM-based classifier needs to be tuned beyond the engineered prompts. 
Since dataset-specific tuning data (paired image-text) is expensive to collect, self-supervised \cite{maxi,lafter} and text-supervised \cite{lafter} VLM fine-tuning approaches were proposed. 
These approaches do not use any paired data and instead successfully leverage unpaired uni-modal data, such as unlabeled image (or video) collections \cite{lafter,maxi} and text-only samples generated from a strong Large Language Model (LLM) \cite{maxi}, effectively tapping on its \textit{visual world knowledge}. 
In particular, LaFTer~\cite{lafter} proposes to substitute visual instances by prompting the LLM to generate textual descriptions of categories in the downstream dataset and uses the generated samples as training data for a VLM-based visual classifier. 

\begin{figure}[t!]
\centering
\centerline{\includegraphics[width=8.0cm]{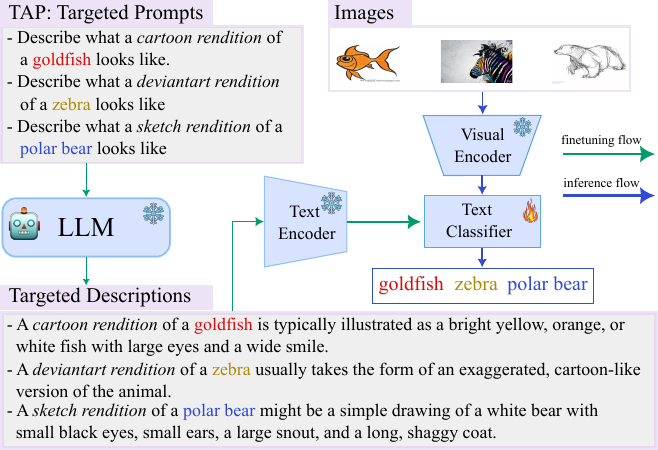}}
\vspace{-0.7em}
\caption{\textbf{Incorporating the visual characteristics of the downstream task with TAP.}
Task-specific LLM prompts provide multiple samples of dataset-specific class descriptions. These are used to train a text-based classifier, tasked with predicting the corresponding class names. Finally, this pre-trained text-only classifier is directly applied to classify visual data, leveraging the shared image-text embedding space learned by the VLMs. At runtime, TAP applies \textit{all} possible targeted prompt templates to \textit{all} target task classes equally. This example illustrates TAP for the ImageNet-Rendition~\cite{imagenet-r} dataset.}
\label{fig:teaser}
\vspace{-2.4em}
\end{figure}

The past approaches~\cite{maxi,lafter}, still employ a somewhat \textit{generalist} strategy for leveraging the uni-modal data, employing the same technique regardless of the nature of the downstream (target) task, and in particular, regardless of the visual properties of the target objects of interest.  
Consequently, while demonstrating strong improvements, these methods are not able to take advantage of the full potential of the visual knowledge possessed by modern-day strong LLMs and image-text-aligned representation spaces of VLMs.
While the content of the unlabeled image collections (as used in \cite{maxi,lafter}) is hard to control, in contrast the generation of the text-only samples 
(textual descriptions of objects in the downstream task) for VLM-based visual classifier training can be effectively manipulated using the knowledge about the visual properties of the downstream (target) task.
In this paper, we demonstrate for the first time a method and empirical evidence that specific downstream targeting is possible while generating these text-only training samples. 
Specifically, we show significant performance improvements when incorporating the textual descriptions of the visual characteristics of the downstream tasks into the prompts for the LLM that are used to generate the text-only samples. 
As illustrated in Figure~\ref{fig:teaser}, these generated samples are in turn used to learn a text classifier (trained to classify textual descriptions to class names), which can be readily applied to classify the visual data, foregoing the constraint to obtain paired image-text data.

In this work, we identify several types of domain shifts (from the large-scale VLM, e.g., CLIP pre-training) arising while transferring the VLM-based zero-shot visual classifier to downstream applications, and propose a simple principled approach for generating the text-only samples for each type.
Our key idea is~\textbf{Ta}rgeted-\textbf{P}rompting (TAP): prompting the LLM to explicitly specify the desired visual characteristics of objects to highlight while generating the text-only training samples.
TAP leverages the \emph{visual world knowledge} and \emph{domain expertise} acquired by LLMs during their large-scale pre-training on trillions-of-tokens-scale internet data. 
The samples generated via our TAP exhibit greater visual detail in aspects relevant to the expected domain shift corresponding to the target downstream task and consequently lead to greater improvement when used for finetuning.
Our simple targeted prompting strategy results in significant performance gains of the visual VLM-based classifier compared to a collection of strong baselines including the base VLM, its adaptation and SOTA prior work (\cite{lafter}) for text-only VLM finetuning.

\vspace{-1em}
\section{Related work}
\vspace{-1em}

VLMs pre-trained on large-scale web data (e.g., CLIP~\cite{clip}) show strong zero-shot performance for recognition tasks. 
Recently, many interesting extensions to improve the image-text representation alignment have been proposed
~\cite{albef,tcl,blip, furst2021cloob,declip,gao2021clip}. 
Parallel works have explored VLM improvements with: extra supervision~\cite{li2021supervision,mu2021slip}, fine-grained interactions~\cite{yao2021filip}, modern Hopfield networks~\cite{furst2021cloob}, optimal transport distillation~\cite{wu2021data}, cycle consistency~\cite{goel2022cyclip}, and hierarchical feature alignment~\cite{gao2022pyramidclip}.
Recent works, such as~\cite{coop,cocoop, maple} demonstrate that despite the great progress in zero-shot recognition tasks, VLM performance on downstream tasks can be significantly improved with supervised fine-tuning on annotated data. 
On the other hand, UPL~\cite{upl}, CLIP-PR~\cite{clippr}, MAXI~\cite{maxi}, and LaFTer~\cite{lafter} show that improvements are also possible with finetuning using unlabeled data.
UPL finetunes learnable text prompts (similar to~\cite{coop}) by relying on confidence sampled pseudo-labeling, whereas CLIP-PR relies both on offline pseudo-labeling and label distribution prior from the training data. 
MAXI focuses only on zero-shot action recognition tuning on unlabeled video data and rich text dictionaries for matching and pseudo-labeling. 
LaFTer combines tuning on text samples generated from LLMs using generic (non-task-specific) prompts and pseudo-labeled (online FixMatch \cite{fixmatch} style) image collection.
In our work, we show that not only the text-only LLM-generated samples can be used to tune VLM-based visual classifiers (in contrast to~\cite{upl,clippr,maxi}), but also that (in contrast to LaFTer~\cite{lafter}) those text-samples can be manipulated in a principled manner to become downstream task-adaptive, leading to significant performance improvements (e.g.,~over~\cite{lafter}). 
Extensive empirical evaluations show the benefits of our approach w.r.t. all the leading label-free finetuning baselines.

\vspace{-1em}
\section{TAP: Targeted Prompting}
\vspace{-1em}
CLIP~\cite{clip} is trained on a large corpus of paired text and image data collected from various sources on the internet. It consists of visual and text encoders that project textual and visual data into a high-dimensional space. 
The encoders are trained using a contrastive loss, which encourages embeddings of same text-image pairs to be closer in the embedding space while pushing apart embeddings of texts and images belonging to different pairs. 
The shared text-image embedding space learned by CLIP (due to the contrastive loss) enables effective zero-shot image classification.
To achieve this, the test image $x$ is encoded by the vision encoder $v$ to a latent space. 
Similarly, the text encoder $\txtEnc$ is used to generate the embedding $\txtEmb_\cls$ for a text prompt for each class $\cls\in\ClassNames$, where $\ClassNames$ represent the set of class names for the respective downstream dataset. 
The prompts can be of the form `\emph{A photo of a \{\cls\}}'.
The zero-shot classification is performed by calculating the cosine similarity of the visual embedding with the text embeddings of each class. 
More formally, the probability for a predicted class~$\hat\cls$ is computed with the temperature-scaled softmax,
\begin{equation} \label{eq:zeroshot}
\likel_{\hat\cls}(\img) = \frac{e^{\simil(\txtEmb_{\hat\cls},\visEnc(\img))/{\tau}}}{\sum_{\cls\in\ClassNames} e^{\simil(\txtEmb_{\cls},\visEnc(\img))/{\tau}}}  .
\end{equation}

In~\cite{clip} it was highlighted that obtaining the text embeddings by using different prompting strategies can have an effect on the eventual zero-shot classification performance.
Recent works, such as CuPL~\cite{cupl} and LaFTer~\cite{lafter} show that by generating the textual embeddings from generic descriptions of the categories in the dataset (produced by an LLM in response to a generic prompt, such as \texttt{Describe an image of a \emph{\{category\}}}) can improve the zero-shot classification results. 
In contrast, in this work we propose to replace visual instances by generating text-only class descriptions by prompting an LLM in a \emph{targeted} manner that allows taking the visual characteristics of the downstream task into account, leading to significantly better classification performance than any of the previously proposed (human engineered or LLM produced) generic class prompts. 
In Section~\ref{subsec:targetted_prompts}, we describe our LLM \textbf{TA}rgeted \textbf{P}rompting (TAP) strategy, and in Section~\ref{subsec:cross_modal_transfer} we provide details about how the generated descriptions can be used to replace visual instances by training a text classifier which can effectively classify visual data, leveraging cross-modal transfer capabilities of CLIP.   
\vspace{-1em}
\subsection{Targeted Prompts to LLM}
\vspace{-0.5em}
\label{subsec:targetted_prompts}
Previous works~\cite{cupl,lafter} have highlighted the significance of generating descriptions of categories in a dataset (through an LLM) and using those descriptions (instead of simple templates provided by CLIP~\cite{clip}) to obtain the text embeddings for improved zero-shot classification.
Our TAP also relies on extracting the visual knowledge from the LLM for improved zero-shot classification. 
However, contrary to the previous approaches we propose to prompt the LLM in a \emph{targeted} manner.
While~\cite{cupl,lafter} use the same generic prompts for generating (visual scene) descriptions for improving zero-shot performance 
for all the downstream datasets, in TAP we propose and show for the first time the \textit{surprising effectiveness} of adapting the prompts 
(for generating descriptions)
to the downstream dataset for which we need to perform the zero-shot classification. 
This allows TAP to exploit the additional LLM knowledge about visual properties (characteristics) of class images of the downstream task that do not surface in the output when using generic prompts. 
For example, the specific characteristics of renditions, other imaging modalities (such as satellite imagery), or fine-grained distinctions between very similar classes.

We have identified two principled targeted prompting strategies that make TAP effective. 
These two strategies are focused on addressing the two prevalent domain shifts appearing in many downstream tasks, namely: (i) cross-visual-domain shifts between natural imagery (of VLM web pre-training) to various kinds of renditions or other imaging modalities (e.g., microscopes or satellites), and (ii) fine-grained shifts, when the classifier sensitivity to fine-grained distinctions between similar downstream task classes needs to be enhanced. Surprisingly, we found that simple targeted prompting strategies work best for both of those important cases.

\begin{table}
\setlength\tabcolsep{4pt}
    \centering
    \tiny
    \begin{tabular}{c|l|l}
    \toprule
        &  \multicolumn{1}{c|}{\textbf{Generic Prompting~\cite{cupl, lafter}}} &\multicolumn{1}{c}{\textbf{Targeted Prompting}}\\
         \midrule
         \midrule
        {\multirow{6}{*}{\rotatebox[origin=c]{90}{\textbf{DTD}}}}  & \llmprompt{P:} Describe what a \class{banded} looks like.
         & \llmprompt{P:} Describe what a \class{banded} \targetclass{texture} looks like.\\
         & \llmdescription{D:} Banded snakes have yellow or white stripes
                  & \llmdescription{D:} A banded texture has distinct lines or stripes of\\
         & \phantom{\textbf{P:}  }wrapping around their body.
         &\phanthomd{}contrasting colors within same color family.\\
\cmidrule{2-3}         
& \llmprompt{P:} How can you identify a \class{braided}.
         &\llmprompt{P:} Describe what a \class{braided} \targetclass{texture} looks like.\\
         &\llmdescription{D:} A braided hairstyle can be identified by its
                  &\llmdescription{D:} Braided texture can be identified by its\\
         &\phanthomd{}interwoven sections of hair.
         &\phanthomd{}woven-like appearance and feel.\\

         \midrule
         {\multirow{6}{*}{\rotatebox[origin=c]{90}{\textbf{Flowers-102}}}} &\llmprompt{P:} Describe what an \class{artichoke} looks like.
         &\llmprompt{P:} Describe what an \class{artichoke} \targetclass{flower} looks like.\\
         &\llmdescription{D:} An artichoke is a large, round, and thistle-like
                  &\llmdescription{D:} An artichoke is a large flower with a spiny\\
         &\phanthomd{}vegetable.
         &\phanthomd{}shape and thick, tough leaves.\\
\cmidrule{2-3}         
&\llmprompt{P:} How can you identify a \class{fritillary}?
         &\llmprompt{P:} How can you identify a \class{fritillary} \targetclass{flower}?\\
         &\llmdescription{D:} Fritillary butterfly is identified by its checked
                  &\llmdescription{D:} Fritillary has distinctive bell or pendulous\\
         &\phanthomd{}pattern on the upper side of their wings.
         &\phanthomd{}shape, usually with six petals.\\

         \midrule
         {\multirow{6}{*}{\rotatebox[origin=c]{90}{\textbf{EuroSat}}}} & \llmprompt{P:} Describe what a \class{forest} looks like.
         &\llmprompt{P:} Describe what a \class{forest} looks like \targetclass{from a satellite}.\\
         &\llmdescription{D:} A forest is an area of land covered with trees,
                  &\llmdescription{D:} From a satellite, a forest is likely to show a\\
         &\phanthomd{}bushes, and other vegetation.
         &\phanthomd{}patchwork of varying shades of green.\\
\cmidrule{2-3}         
&\llmprompt{P:} How can you identify a \class{river}?
         &\llmprompt{P:} How can you identify a \class{river} \targetclass{from a satellite}?\\
         &\llmdescription{D:} Rivers can be identified by looking for long,
                  &\llmdescription{D:} To identify a river from a satellite, look for\\
         &\phanthomd{}interwoven sections of hair.
         &\phanthomd{}features like large meanders and wide curves.\\
         \bottomrule
         \bottomrule
    \end{tabular}
        \vspace{-1.3em}
    \caption{\textbf{Comparing \textit{generic} and \textit{TAP} prompts.} Showing two example LLM \llmprompt{prompts} and resulting \llmdescription{descriptions} (from GPT-3.5, DaVinci-003) for two \class{classes} of Describable Textures (DTD)~\cite{dtd}, Flowers-102~\cite{flowers102} and EuroSat~\cite{eurosat} datasets.}
    \label{tab:generic_targeted_prompts}
    \vspace{-0.7cm}
\end{table}

For fine-grained shifts, we discovered that providing super-class context is of great importance for the LLM to produce better fitting textual description instances.
Table~\ref{tab:generic_targeted_prompts} compares the generic prompting adopted by~\cite{cupl, lafter} with our Targeted Prompting strategy.
Using generic prompts, it is difficult for the LLM to relate fine-grained categories to their super-class.
For example, for \emph{texture} classification on DTD~\cite{dtd}, the LLM confuses \texttt{banded} as a type of \emph{snake}.
Similarly, without the super-class context, the LLM wrongly describes the~\texttt{braided} texture as a \emph{hairstyle}. 
Due to the same word used in the English language for categories belonging to totally un-related super-classes, the LLM provides erroneous descriptions for categories in Flowers-102~\cite{flowers102} dataset as well. 
For example, the generic prompting results in the description of the~\texttt{fritillary}~\emph{butterfly} instead of the~\emph{flower}.
To obtain descriptions correlated with the downstream task, our targeted prompts provide the LLM with the additional context that it needs to describe~\texttt{braided} as a form of~\emph{texture} and~\texttt{fritillary} as a form of~\emph{flower}.

Our strategy for cross-visual-domain shifts is illustrated in Figure~\ref{fig:teaser} for ImageNet-R~\cite{imagenet-r} and in Table~\ref{tab:generic_targeted_prompts} for EuroSat~\cite{eurosat}. 
We have found that LLMs generate effective target domain descriptions if explicitly prompted via short descriptions of the visual characteristics of the domain of interest. 
For example, for rendition domains, we can specify types of renditions of interest, requesting an LLM to produce description instances for a Cartesian product between the set of classes and the set of rendition types. For satellite or microscopy imagery we specifically request descriptions in those imaging domains. For action videos (UCF-101~\cite{ucf101}), we ask the LLM to highlight the temporal and motion aspects prompting for \textit{action} descriptions. For, scenery (SUN-397~\cite{sun397}) we ask the LLM to focus the background features of the visual scene asking for \textit{scene} descriptions.
All targeted prompts and descriptions for the different types of domain shifts are included in our complete codebase available at \href{https://github.com/jmiemirza/TAP}{https://github.com/jmiemirza/TAP}.
\vspace{-1.em}
\subsection{Cross-Modal Transfer}
\vspace{-0.5em}
\label{subsec:cross_modal_transfer}
The Vision Language models have an aligned text and image embedding space due to the contrastive learning objective between the images and the alt-text pairs, used for pre-training.
This allows these models to be used as an effective zero-shot classifier.
Due to the shared embedding space learned by these multi-modal models, recently, DrML~\cite{zhang2022drml} shows that the behavior of visual classifiers can be rectified by using textual data only. 
Furthermore, LaFTer~\cite{lafter} highlights another advantage of the shared embedding space, where it firsts train a classifier to classify (multiple samples of) LLM-generated text descriptions of classes to the correct class names in the dataset and then applies the trained classifier to classify visual embeddings, highlighting effective cross-modal transfer.  
\begin{table*}[t!]
    \centering
    \scriptsize
    \begin{tabular}{lcccccccc|c}
    \toprule
         &\textbf{Describable Textures}& \textbf{Flowers-102} &\textbf{ImageNet-Rendition} &\textbf{FGVC-Aircraft}&\textbf{UCF-101}& \textbf{Food-101} &\textbf{EuroSat} &\textbf{SUN-397}&\textbf{Mean}\\
         \midrule
        \midrule
         CLIP (single)&40.3&	64.0	&65.8	&18.1&61.0&	77.1	&35.9&	60.8&52.8\\
         TOT (cls-only)&37.4&60.4&63.1&15.4&56.4&74.3&39.1&56.3&50.3\\
         CLIP (DST) &42.4&\underline{66.6}&\underline{68.6}&\underline{19.5}&62.4& \textbf{79.3}&\underline{45.8}&\underline{61.7}&\underline{55.8}\\
         TOT (DST)&41.6&64.5&67.5&17.2&62.0&77.2&43.6&61.0&54.3\\
         LaFTer (text-cls)&\underline{44.3}&64.2&63.6&18.3&\underline{63.7}&78.1&44.2&60.7&54.6\\
         \midrule
         TAP&{\textbf{51.1}}&\textbf{{66.9}}&\textbf{68.9}&\textbf{21.6}&{\textbf{66.3}}&\underline{78.5}&\textbf{54.2}&\textbf{63.4}&\textbf{58.9}\\
         \bottomrule
         \bottomrule
    \end{tabular}
    \vspace{-0.8em}
    \caption{
    Top-1 Classification Accuracy (\%) for a ViT-B/32 CLIP pre-trained network. \textit{CLIP} denotes the zero-shot evaluation performed by using either \textit{single} prompt template or a set of \textit{dataset-specific prompt templates} (DST). Text-Only Training (TOT) results are obtained by training a text classifier to classify either the class names only or the DST texts.
    For a fair comparison, we also compare with results from text-only pre-trained visual classifier in LaFTer~\cite{lafter}. The highest accuracy is shown in \textbf{bold}, while the second best is \underline{underlined}. }
    \label{tab:text_only_results}
    \vspace{-1.5em}
\end{table*}

\begin{table}[t]
\setlength\tabcolsep{5pt}

    \centering
    \scriptsize
    \begin{tabular}{lcccc|c}
    \toprule
         &  \textbf{Flowers-102}&\textbf{UCF-101}&\textbf{DTD}&\textbf{IN-R}&\textbf{Mean}\\
         \midrule
         \midrule
         CLIP& 66.6 &62.4&42.4&68.6&60.0\\
         CLIP-PR&60.1&59.7&45.1&56.9&55.5\\
         UPL&\underline{71.5}&63.9&\underline{48.0}&65.6&62.3\\
         LaFTer& {71.0}&\underline{68.2}&{46.1}&\underline{72.6}&\underline{64.5}\\ 
         \midrule
            TAP$^*$&\textbf{72.4}&\textbf{71.4}&\textbf{51.6}&\textbf{73.9}&\textbf{67.3}\\
         \bottomrule
         \bottomrule
    \end{tabular}
        \vspace{-0.8em}

    \caption{Top-1 Classification Accuracy (\%). TAP$^*$ result is obtained by further finetuning the vision encoder of CLIP, with pseudo-labels generated for an unlabeled image collection by the TAP text-classifier, following the complete LaFTer~\cite{lafter} pipeline. }
    \label{tab:full_lafter}
    \vspace{-2em}
\end{table} 
Cross-Modality transfer introduces an efficient way to train neural networks on relatively cheap text data and avoid the manual labor of acquiring and annotating images. 
A large quantity of text data can be conveniently harnessed by prompting the LLM for descriptions of classes, which can effectively replace visual instances. 
This is also a relatively easier alternative to mining text data, since LLMs already represent a large corpus of text on which they were trained. 
In this work, we take inspiration from~\cite{lafter} and also train a classifier on a textual dataset constructed by generating descriptions of classes.
However, contrary to LaFTer~\cite{lafter}, we generate descriptions for the downstream dataset by explicitly prompting the LLM in a targeted manner as described in Section~\ref{subsec:targetted_prompts}.
Moreover, unlike LaFTer, our TAP does not rely on generating the text data by mixing hand-crafted templates from~\cite{clip} with the text descriptions, instead, we use the raw descriptions generated by the LLM without requiring any post-processing.
\begin{figure}
    \centering
    \includegraphics{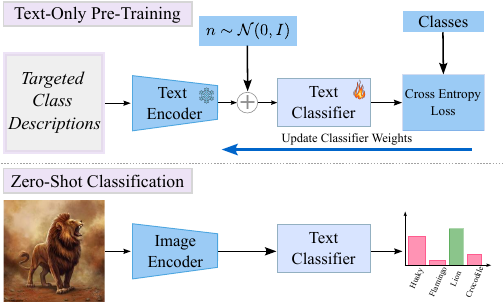}
    \vspace{-0.45em}
    \caption{The targeted descriptions generated by the LLM are automatically matched with the class names to generate the text dataset. This dataset is used to train a text classifier in a \emph{supervised} setting (top). The trained text classifier is used to classify the visual data (bottom).}
    \vspace{-1.8em}
    \label{fig:methodology}
\end{figure}

The class descriptions from the LLM are automatically matched with the class name and training the text classifier can be rendered as a \textit{supervised} learning problem, as depicted in Figure~\ref{fig:methodology}. 
Applying the trained text classifier to classify visual embeddings also allows us to forego constraints that are being imposed by the classical (VLM) zero-shot classification setup, according to Eq.~\eqref{eq:zeroshot}.
In contrast, for TAP, to obtain the class likelihoods for the visual classification, we can directly employ the trained text classifier~$f$ instead of computing the cosine similarity between embeddings from different modalities. 
More formally, given a text dataset $T$, consisting of class descriptions $t$ and their corresponding class labels $\cls$, the text classifier is trained with the following \emph{supervised} training objective:
\begin{equation} \label{eq:txt-pretraining}
\min_\theta
\sum_{\substack{(t,\cls)\in T\\ 
                n \sim \mathcal{N}(0,I)}}
\LSCE \big(f_\theta (\frac{\txtEnc(t)}{\lVert\txtEnc(t)\rVert}+n),\cls\big) ,
\end{equation}
where $\theta$ is the parameter of the classifier, $\LSCE$ represents the Smoothened Cross Entropy loss~\cite{labelsmoothing} and Gaussian Noise $n \sim \mathcal{N}(0, I)$ is added to the text embeddings to regularize training. 

\vspace{-1em}
\section{Experimental Evaluation}
\vspace{-1em}
\subsection{Evaluation Details}
\vspace{-0.5em}
\paragraph{Datasets} We extensively evaluate our TAP on $8$ different datasets which include several fine-grained classification datasets, such as Flowers-102~\cite{flowers102}, FGVC Aircraft~\cite{aircraft}, Food-101~\cite{food} and Describable Textures Dataset (DTD)~\cite{dtd}, and also domain-specific classification datasets such as UCF-101~\cite{ucf101}, EuroSat~\cite{eurosat}, SUN-397~\cite{sun397} and ImageNet-Rendition (IN-R)~\cite{imagenet-r}. 
These groups exhibit the two prominent domain shifts from VLM (e.g., CLIP \cite{clip}) web-data pre-training to downstream zero-shot application, as discussed in Section~\ref{subsec:targetted_prompts}.

\paragraph{Baselines} We compare our method with several strong baselines: 
\begin{itemize}[leftmargin=0pt]
    \item \textbf{CLIP~\cite{clip}:\flexquad} represents the zero-shot evaluations performed by using the base CLIP network. 
    We evaluate CLIP by using a \textit{single} prompt template~`\emph{a photo of a \{c\}}' and also the full set of \textit{dataset-specific templates} (denoted DST) provided by~\cite{clip} for a fair evaluation. 
    \item \textbf{Text-Only Training (TOT):\flexquad} represents the results obtained by training a classifier (similar to our TAP), but with two different training objectives -- classifying the class name; TOT (cls-only), and classifying DST to the class names; TOT (DST).  
    \item \textbf{LaFTer~\cite{lafter}:\flexquad} trains a text classifier on LLM descriptions and dataset-specific templates from~\cite{clip} and later employs unsupervised fine-tuning of the visual encoder via pseudo-labeled visual data.

\end{itemize}

\paragraph{Implementation Details}
For all experiments, we use the OpenAI pre-trained ViT-B/32 CLIP~\cite{clip} model.
To ensure a fair comparison with~\cite{lafter}, the text classifier (Section~\ref{subsec:cross_modal_transfer}) is implemented as a single fully connected layer, where the number of the output units of the layer equals the number of classes in the downstream dataset.
Similarly, following~\cite{lafter} we use GPT-3.5 DaVinci-003 to generate the descriptions.
The text dataset is loaded as a single batch and the classifier is optimized using AdamW, with a learning rate of $0.001$. 
All details can be found in our codebase: \href{https://github.com/jmiemirza/TAP}{https://github.com/jmiemirza/TAP}

\vspace{-1em}
\subsection{Results}
\vspace{-0.5em}
\label{sec:refs}
Table~\ref{tab:text_only_results} shows the comparison against (unadapted) CLIP and baselines which do not require visual instances to finetune the VLM. 
Our TAP consistently improves over the zero-shot CLIP baseline (with a single `\textit{a photo of a ...}' template) on all the $8$ datasets, showing up to $18.3\%$ improvement on EuroSat ($6.1\%$ on average).
On IN-R our TAP improves the base CLIP by $3.3\%$, which is an out-of-distribution variant of the original ImageNet~\cite{imagenet} dataset. 
This improvement highlights that generating text descriptions can be helpful in mitigating the distribution shifts from the text side, which are automatically transferred to the vision domain due to the shared embedding space. 

We also compare our TAP with a stronger CLIP baseline, where text embeddings are generated by using a collection of dataset-specific templates, also referred to as \emph{prompt ensembling} in~\cite{clip}. 
Our TAP still shows $3.1\%$ average gains over the $8$ datasets.
Individually, our TAP improves over $7$ out of $8$ datasets, while remaining competitive on Food-101, where only $0.3\%$ degradation is observed. 
These results highlight the benefits of cross-modal transfer through TAP, where training a text-classifier (on texts automatically generated by an LLM, without requiring any labeled image data) to classify visual data can be a better alternative for zero-shot classification than obtaining the class likelihoods through cosine similarity. 

Our TAP also performs favorably when compared with methods that train a text-classifier on text datasets generated through different methods. 
For example, while training the classifier to only classify the class names or dataset-specific templates to the class names, TOT (cls-only) and TOT (DST) in Table \ref{tab:text_only_results}, our TAP shows an average gain of $8.6\%$ and $4.6\%$.
Moreover, while comparing with text-only pre-training of a classifier through LaFTer, where the text dataset is constructed by combining dataset-specific templates and (generic) LLM descriptions of classes, we show an average improvement of $4.3\%$. 
The gains over LaFTer further highlights the importance of generating targeted descriptions with targeted prompting (TAP) as compared to generating descriptions with a generic prompting technique (as used in LaFTer).  
\vspace{-1em}
\subsection{Ablation Study}
\vspace{-0.5em}
In their second stage, LaFTer~\cite{lafter} also leverages the unlabeled visual instances and finetunes the visual encoder from CLIP by using the pseudo-labels generated by the text-only pre-trained visual classifier (inspired from FixMatch~\cite{fixmatch}). 
The quality of the generated pseudo-labels depends on the pre-trained text classifier.
Thanks to the targeted descriptions from the LLM, since our TAP's text-only classifier provides better results than the LaFTer's text-only pre-training (Table~\ref{tab:text_only_results}), TAP can be employed seamlessly in the second stage of LaFTer~(hereby denoted as TAP$^*$) to provide a further improvement over the full LaFTer pipeline. 
In Table~\ref{tab:full_lafter} we compare TAP$^*$ with LaFTer~\cite{lafter}, 
CLIP-PR~\cite{clippr}, and UPL~\cite{upl} - all of which use unlabeled image data for unsupervised finetuning. Compared to the leading baselines (LaFTer / UPL) TAP$^*$ obtains up to $3.6\%$ gain on the DTD dataset, while on average being $7.3\%$ and $2.8\%$ better than CLIP and the best baseline (LaFTer) on the $4$ datasets. 
\vspace{-1em}
\section{Conclusion and Discussion}
\vspace{-1em}
We have introduced TAP - a simple and effective approach for leveraging LLM's detailed visual world knowledge to improve zero-shot classification in VLMs. 
Our approach only requires knowledge of the class names and the general visual characteristics of the target downstream task to make the VLM-based visual classifier more effective in zero-shot recognition on the target task when compared to a variety of baselines.
Our work may lead to a number of exciting follow-ups; currently, TAP requires some basic human knowledge of visual characteristics in the downstream task to prompt the LLM to generate targeted textual scene instances, later used for text-only pre-training.
As future work, it might also be possible to obtain these visual characteristics by automatically prompting the LLM itself, effectively creating a self-contained and almost no-human in-the-loop two-stage approach.  
TAP can also be naturally extended to mitigate other domain shifts, 
such as:
adversarial and outlier robustness, long tail category distributions, unbalanced class hierarchies, and more.
{
\small
\bibliographystyle{IEEEbib}
\bibliography{icassp}
}
\end{document}